# Adaptive Group Collaborative Artificial Bee Colony Algorithm

Haiquan Wang,[1] Hans-Dietrich Haasis,[2] Panpan Du,[3] Xiaobin Xu,[4] Menghao Su,[3] Shengjun Wen,[1] Wenxuan Yue,[3] and Shanshan Zhang[3]

[1] Zhongyuan Petersburg Aviation College, Zhongyuan University of Technology, Zhengzhou, 450007, China
[2] Business Studies and Economics, University of Bremen, Zinckestraße 20, 28865 Lilienthal, Germany
[3] Faculty of Electrical and Engineering, Zhongyuan University of Technology, Zhengzhou, 450007, China
[4] School of Automation, Hangzhou Dianzi University, 115 Wenyi Road, Xihu District, Hangzhou 310018, China

Correspondence should be addressed to Haiquan Wang; wanghq@zut.edu.cn

## Abstract

As an effective algorithm for solving complex optimization problems, artificial bee colony (ABC) algorithm has shown to be competitive, but the same as other population-based algorithms, it is poor at balancing the abilities of global searching in the whole solution space (named as exploration) and quick searching in local solution space which is defined as exploitation. For improving the performance of ABC, an adaptive group collaborative ABC (AgABC) algorithm is introduced where the population in different phases is divided to specific groups and different search strategies with different abilities are assigned to the members in groups, and the member or strategy which obtains the best solution will be employed for further searching. Experimental results on benchmark functions show that the proposed algorithm with dynamic mechanism is superior to other algorithms in searching accuracy and stability. Furthermore, numerical experiments show that the proposed method can generate the optimal solution for the complex scheduling problem.

## Introduction

The artificial bee colony algorithm was firstly proposed by Karaboga 0. It has been widely used in power system scheduling, architecture design, storage system optimization, etc, as it possesses the global optimization ability 0. However, it also suffers problems such as poor at exploitation and easily falling into local optimum, which are the common failings of all intelligent algorithms. For improving its performance, many researches focus on introducing other algorithms and combining their updating mechanisms with ABC, Devi 0 proposed an improved artificial bee colony firefly algorithm (MABCFA), where the attraction model of the firefly algorithm is added to the neighbourhood search process to improve the algorithm's convergence performance. Rambabu et al. 0 incorporated the mutated butterfly adjusting operator of Mutated Butterfly Optimization Algorithm (MBOA) with the employee bee phase of ABC for preventing earlier trapping into local optimal. Liang et al. 0 introduced differential operator in the employed bee phase to produce candidate solutions with increasing probability. The firework explosion search mechanism was combined with ABC where the fireworks explosion search stage was added to search around the potential food sources in 0. Besides these improvements, two or more search strategies were introduced to replace the single search strategy in original ABC. Literature 0 proposed two search strategies based on the best-so-far solution information and the means of two random solutions. Xu et al. 0 introduced the differential evolution to the employed bee phase to accelerate its convergence and adopted the global best position to guide the evolution direction of onlooker bees. In 0, three search strategies with different characteristics were employed to construct a strategy candidate pool and the Parzen window method was applied to estimate and select high-quality candidate individuals. However, the problem is that they are impossible to judge whether the proposed search strategies are fit for specific optimization problems, and different strategies should be adjusted by trial and error during the optimization procedure. In this paper, another improvement, where the update process is guided by the group collaboration mechanism, is introduced and different searching strategies with different characteristics and adaptive hybrid proportions are used to balance the ability of exploration and exploitation of the algorithm.

The rest of this paper is organized as follows: The improved ABC algorithm is proposed in Section 2, and its performance is verified with benchmark functions. In Section 3, the algorithm is applied to the twin ETVs scheduling problem with constraints. Finally, the above work is summarized.

## The improved ABC algorithm

**ABC algorithm**

ABC algorithm implements a process of iterative optimization, during which the bee colony is divided into three groups: employed bees, onlooker bees and scout bees[11]. The algorithm starts from employed bee phase and executes crossover and mutation process with one randomly chosen companion, thus the new solution $x'_{mk}$ could be updated based on $x_{mk}$ as Equation 1. Then the fitness value of each solution *fitness$_m$* which represents its quality could be calculated, and the onlooker bee chooses to exploit or not around corresponding employed bee with the probability *Pm* defined as Equation 2. Here the *kth* companion is still randomly selected. If the current *mth* solution to be exploited can't be improved for several iterations, the solution will be abandoned, and a scout bee will replace it with a new randomly produced solution.

$$x'_{mk} = x_{mk} + rand(-1,1) * (x_{mk} - x_{nk}) \qquad (1)$$

$$P_m = \frac{fitness_m}{\sum_{m=1}^{SN} fitness_m} \qquad (2)$$

Where *m*, *n* represent the indices of specific solution in the population, $m,n \in \{1,2,…,N\}$, $m \neq n$. $k \in \{1,2,…,D\}$ represents the dimension of the population. $rand(-1,1)$ is a random number between [-1,1]. *SN* is the number of food sources which is equal to the number of employed bees.

**The improved ABC algorithm**

In ABC algorithm, only one randomly searching strategy as Equation 1 is adopted in the employed bee and onlooker bee phases. For making the algorithm more flexible and adaptable for all backgrounds, an adaptive group collaborative mechanism, which applies several searching strategies with different functions to different phases or different populations, is designed.

**The improved searching strategies**

Several searching strategies with different aims, such as the updated strategy with Levy flight mechanism which could expand the search area as well as the strategy with variable step coefficient which could execute local accurate search, are proposed.

(1) Long step search strategy

In order to increase the step length during the evolution process, Levy flight mechanism 0 which mimics the migration of birds in nature or builds nests and realizes long-distance migration behaviour, is introduced to strengthen the ability of exploration in solution space. The first improvement as shown in Equation 3 is realized with the help of Levy flight operator during the neighbourhood search process, and the second one described as Equation 4 substitutes the traditional random search strategy with the levy flight mechanism.

$$x'_{ik} = x_{ik} + rand(-1,1) \times [x_{ik} - levy(x_{jk})] \quad (3)$$

$$x'_{ik} = levy(x_{ik}) \quad (4)$$

Which *levy* represents the Levi flight mechanism, and its updating formula is defined as Equation 5 and 6:

$$levy(x_k) = x_k + \partial \oplus L(\lambda)(i=1,2,...,n) \quad (5)$$

$$L(\lambda) = \left| \frac{\Gamma(1+\lambda) \times \sin(\frac{\pi\lambda}{2})}{\Gamma(\frac{1+\lambda}{2}) \times \lambda \times 2^{(\frac{\lambda-1}{2})}} \right|^{\frac{1}{\lambda}} \quad (6)$$

$x_k$ is the solution that needs to be updated; $\partial$ is the step control variable; $\oplus$ means dot product; $L(\lambda)$ is the Levi flight model with constant $\lambda$; $\Gamma$ is the standard Gamma function; $i$ and $j$ are the population size, $i, j \in \{1,2,...,N\}$ and $i \neq j$, $k \in \{1,2,...,D\}$ is the dimension of solution, $rand(-1,1)$ is the random number between [-1, 1].

With the help of Levy flight mechanism, the search area corresponding to the above search strategies could be expanded.

(2) Short step search strategy

On the other hand, as can be seen from Equation 7, a variable step coefficient $\frac{1}{iter \times W_2}$ is introduced and combined with Equation 1 where *iter* is the iteration number, $W_2$ is the weight which can be set according to specific problem between 0 and 1. Obviously, the step coefficient becomes smaller and smaller as the evolution progresses, and the search gradually concentrates in the local area, thus the local search ability could be enhanced.

$$x'_{ik} = x_{ik} + rand(-1,1) \times (x_{ik} - x_{jk}) \times \frac{1}{iter \times W_2} \quad (7)$$

In order to show the effects of the above proposed strategies in a more intuitive way, evolution procedures for a certain optimal problem are presented in Figures 1-4.

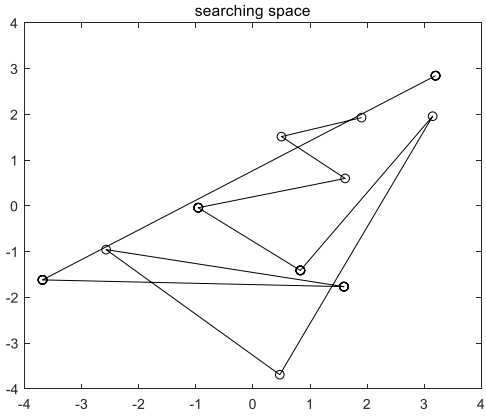

Figure 1: Search Path with Equation 1

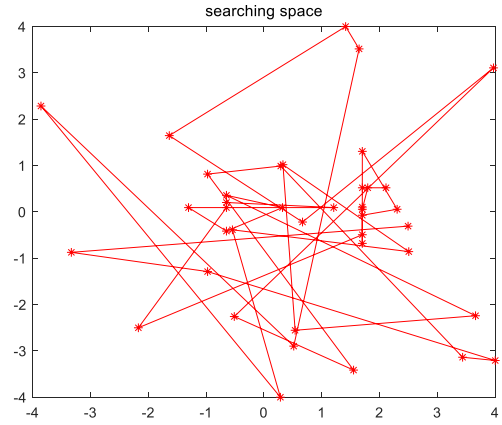

Figure 2: Search path with Equation 4

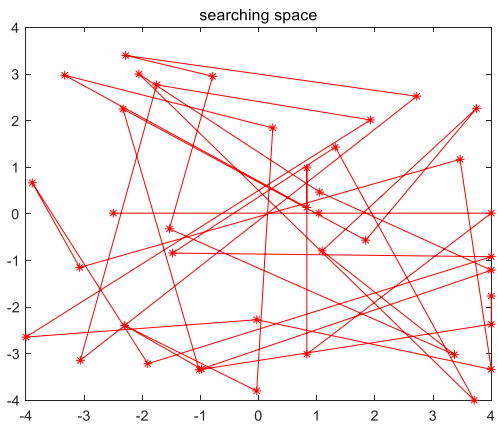

Figure 3: Search path with Equation 3

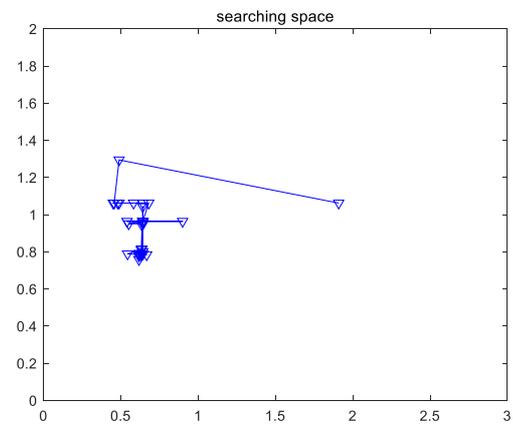

Figure 4: Search path with Equation 7

It can be seen that the search paths of bees are very different corresponding to different search strategies. For long step searching strategies especially the one corresponding to Equation 3, the search paths distribute more widely in the solution space, obviously it is hard to trap into local minimum. And the one corresponding to the short step search strategy converges quickly around the optimal solution, the exploitation ability could be improved.

**The improved adaptive group collaborative mechanism**

Based on the above search strategies, an adaptive group collaborative search mechanism is creatively proposed as follows.

(1) Initialization phase

The employed bees and onlooker bees in the initial population are divided into several groups according to their categories. In this paper, the employed bee group has 3 members in each group, and the onlooker bee group has 2 members. Actually, the number of members in each group can be decided according to the problem to be optimized.

For the members in employed bee groups, long step neighbourhood search strategies as Equations 1, 3, 4, named A, B and C respectively, are used to widen the search. The initial weights of each strategy are the same, and the weights corresponding to the best two search strategies will be added with a constant $W_1$.

For onlooker bee groups, the short-step search strategies as Equation 7 and Equation 8[8], named as D and E, are adopted to improve their convergence capability. They possess the same initial weight values and the increment $W_1$ will be added to the one who possesses the better searching result

$$x'_{ik} = x_{ik} + rand(-1,1) \times (x_{ik} - x_{jk}) + rand(-1,1) \times [G(x_i) - x_{jk}] \times W_3 \qquad (8)$$

$G(x_i)$ is the global optimal solution corresponding to the current iteration number, $W_3$ is the step adjustment factor.

(2) Employed bee phase

Based on the weight values of each strategies, the strategy with the largest weight value has the maximum probability to be selected to execute the following neighbourhood search, the rest of group members or strategies will be kept for next iteration. After updating, the qualities of the best two generated solutions need to be compared with the optimal solution. If its value is better than the optimal solution, the current solution will be stored, and the weight of the corresponding search strategy will be increased by $W_1$.

(3) Onlooker bee phase

Based on the generated weights, different short step search strategies could be selected to perform a neighbourhood search, and the weight of strategy corresponding to the better solution should be further increased.

The pseudo-code of AgABC is given below.

| Adaptive group collaborative ABC |
|---|
| 01  Initialization, set the maximum cycle number, the swarm size, the number of dimensions, the value of *limit* |
| 02      Evaluate the fitness value *fitness(Sol$_m$)* with Equation 2; |
| 03      Add $W_1$ to the best two strategies in each employed bee group(the weights corresponding to the three strategies are $a_1,b_1$ and $c_1$ respectively); |
| 04      Evaluate the fitness value *fitness(Sol$_m$)* with Equation 2; |
| 05      Add $W_1$ to the better strategy in each onlooker bee group(the weights corresponding to the three strategies are $d_1, e_1$ respectively); |
| 06      for *iter* = 1 to Max-cycle number |
|           // Employed bee phase |
| 07          for *m* = 1 to (Number of food sources/2) |
| 08            switch |
| 09              case rand*$a_1$ > rand*$c_1$ && rand*$b_1$ > rand*$c_1$ |
| 10                if *fitness*(A) < *fitness*(Food$_m$) || *fitness*(B) < *fitness*(Food$_m$) |
| 11                   Food$_m$=Sol$_m$; $a_1 = a_1 + W_1$; $b_1 = b_1 + W_1$; sign=1; |
| 12                end if |
| 13              case rand*$a_1$ > rand*$b_1$ && rand*$c_1$ > rand*$b_1$ |
| 14                if *fitness* (B) < *fitness* (Food$_m$) || *fitness* (C) < *fitness* (Food$_m$) |
| 15                   Food$_m$=Sol$_m$; $b_1 = b_1 + W_1$; $c_1 = c_1 + W_1$; sign=1; |
| 16                end if |
| 17              case rand*$b_1$ > rand*$a_1$ && rand*$c_1$ > rand*$a_1$ |
| 18                if *fitness* (A) < *fitness* (Food$_m$) || *fitness* (C) < *fitness* (Food$_m$) |
| 19                   Food$_m$=Sol$_m$, $a_1 = a_1 + W_1$, $c_1 = c_1 + W_1$, sign=1; |
| 20                end if |
| 21            end switch |
| 22            if *sign* = 1 do |
| 23               *trial* = 0; |
| 24            else |
| 25               *trial*=trial +1; |
| 26            end if |
| 27          end for |
|           // onlooker bee phase |
| 28      Calculate the probability *prob$_m$* if the onlooker bees choose to exploit around the specific source |
| 29      for *m*= 1 to Food Number |
| 30         *sign* = 0; |
| 31         if *rand* < *prob$_m$*, do |
| 32            sign=0; |
| 33            if rand*$d_1$> rand*e1 |

```
34              if fitness (D) < fitness (Food_m)
35                 Food_m=Sol_m, d_1 = d_1 + W_1; sign=1;
36              end if
37           else
38              if fitness (E) < fitness (Food_m)
39                 Food_m=Sol_m, e_1 = e_1 + W_1; sign=1;
40              end if
41           end if
42        end if
43        if sign = 1, do
44           trial = 0;
45        else
46           trial=trial +1;
47        end if
48     end for
       // scout bee phase
49     if trial > limit, do
50        trial = 0;
51        randomly generate a new solution
52     end if
53  end for
```

**Performance evaluation**

Seven CEC 2017 benchmark functions as Table 1 are employed to evaluate the performance of AgABC, as well as other algorithms such as ABC, CABC0, GABC[14] and DuABC0. All simulations are executed on an Intel Core i7-8750H CPU with 16G RAM and with a population size of 1000, the dimension of the solution is set to be 60,80 and 100, the number of maximum iterations is set as being 5000, and the limit used in scout bee phase is taken as 100. Independent experiments are run 20 trials with randomly initialized condition, and the fitness value corresponding to the optimal solution (Best-F), average optimal solution (Avg-F), as well as their mean and standard deviation (std) are selected to evaluate the optimization performance of different algorithms.

Table 1: Benchmark functions

| Function expression | Search space | Minimum value | Modality |
|---|---|---|---|
| $f_1(x)=x_1^2+10^6 \sum_{i=2}^{D} x_i^2$ | [-100,100] | 0 | Unimodal |
| $f_2(x) = \sum_{i=1}^{D} (x_i^2 - 10\cos(2\pi x_i)+10)$ | [-100,100] | 0 | Multi-modal |
| $f_3(x) = 0.5 + \frac{\left(\sin\sqrt{x_1^2+x_2^2}\right)^2 - 0.5}{\left(1+0.001(x_1^2+x_2^2)\right)^2}$ | [-100,100] | 0 | Multi-modal |
| $f_4(x) = \sum_{i=1}^{n} |x_i+0.5|^2$ | [-100,100] | 0 | Unimodal |
| $f_5(x) = \sum_{i=2}^{D} x_i^2 + 10^6 x_1^2$ | [-100,100] | 0 | Hybrid |

$$f_6(x) = \sum_{i=1}^{D} \frac{x_i^2}{4000} - \prod_{i=1}^{D} \cos(\frac{x_i}{\sqrt{i}}) + 1 \qquad [-100,100] \quad 0 \quad \text{Hybrid}$$

$$f_7(x) = \left| \sum_{i=1}^{D} x_i^2 - D \right|^{1/4} + (0.5 \sum_{i=1}^{D} x_i^2 + \sum_{i=1}^{D} x_i) / D + 0.5 \qquad [-100,100] \quad 0 \quad \text{Hybrid}$$

The changes of weights corresponding to different strategies during the optimization procedure with AgABC are shown as Figure 5. It can be seen that for function $f_1(x)$, $f_2(x)$, $f_4(x)$, $f_5(x)$, and $f_7(x)$, bees are more inclined to choose strategies B and E, for $f_3(x)$ and $f_6(x)$, bees are more inclined to choose strategies C and E.

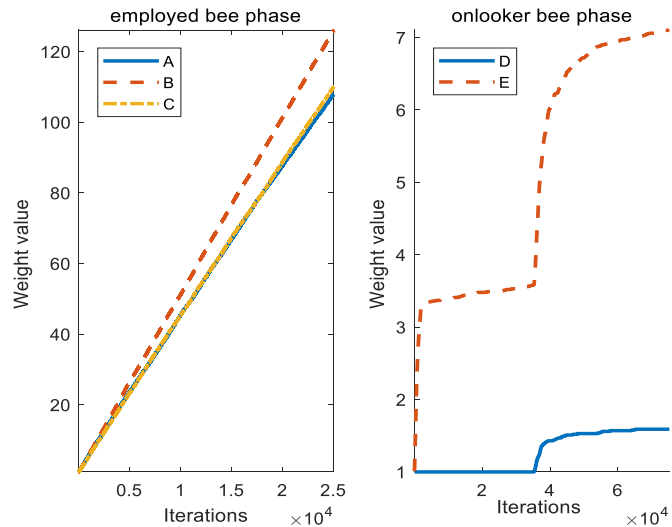

(a) function $f_1(x)$

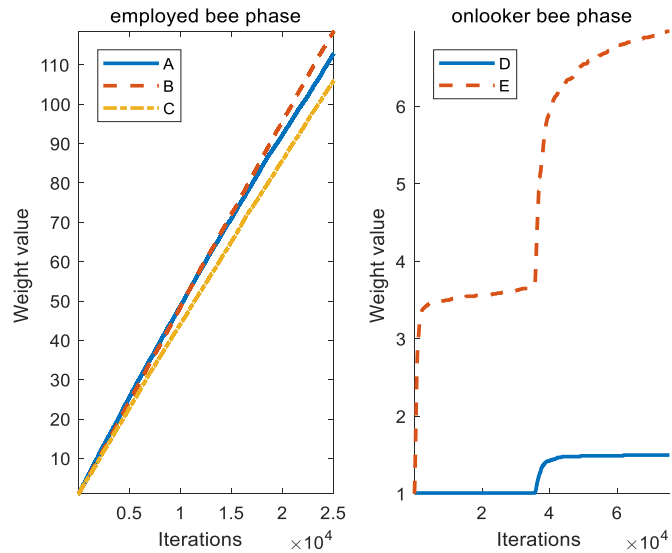

(b) function $f_2(x)$

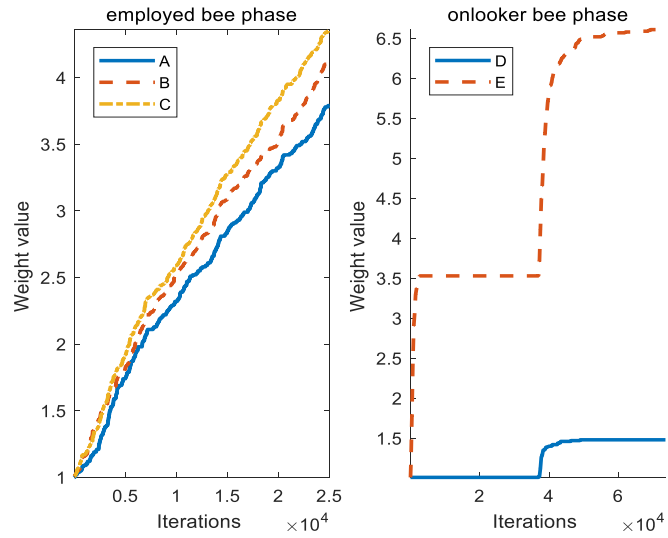

(c) function $f_3(x)$

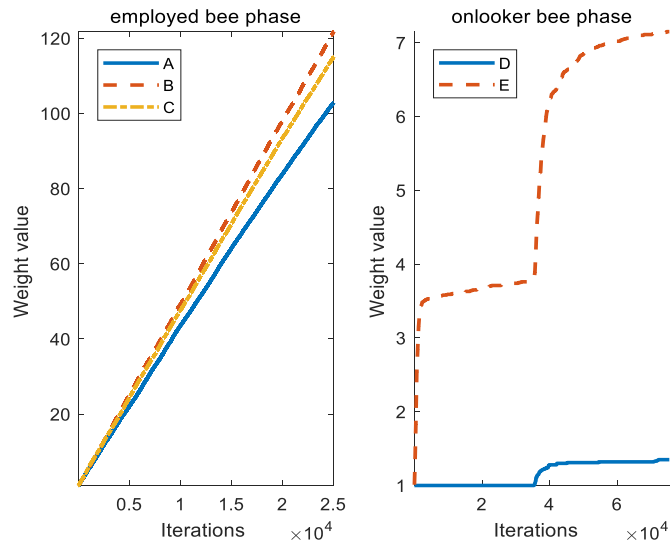

(d) function $f_4(x)$

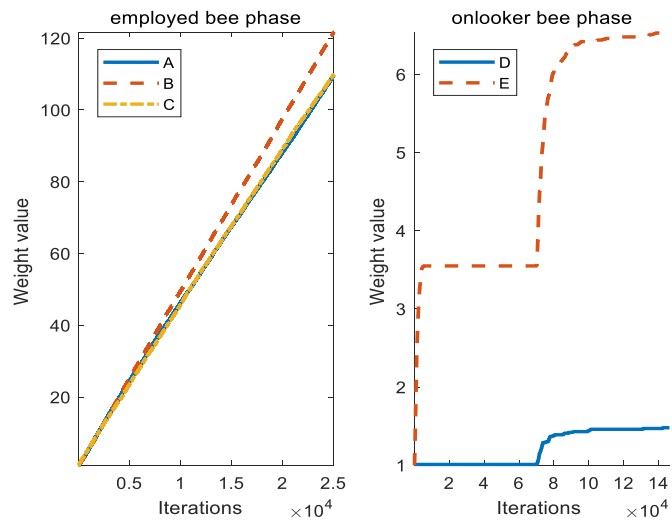

(e) function $f_5(x)$

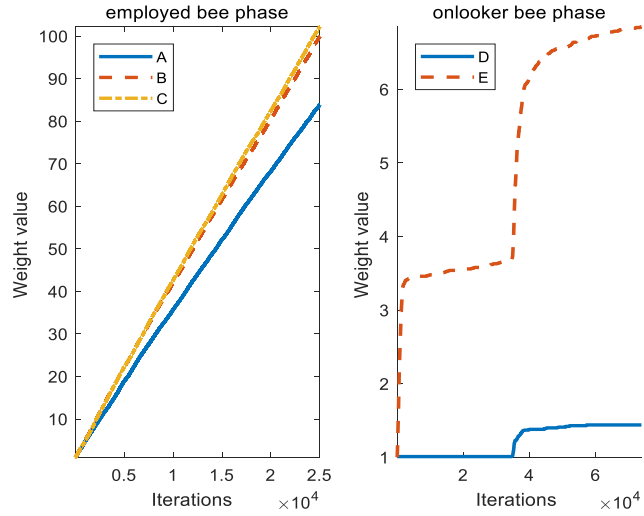

(f) function $f_6(x)$

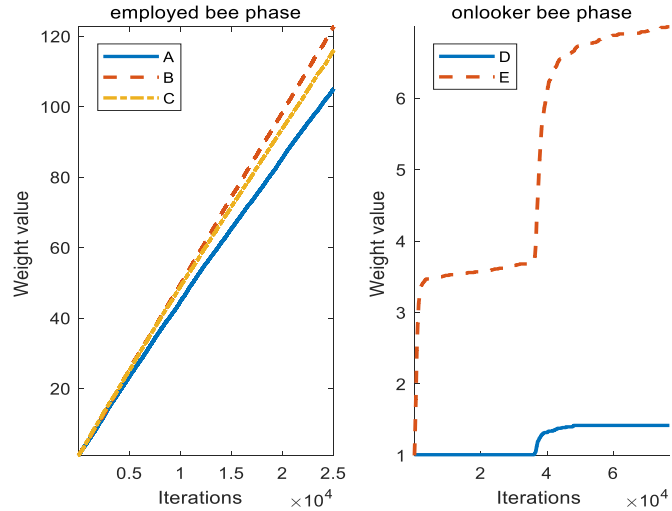

(g) function $f_7(x)$

Figure 5: Change of searching strategy weights corresponding to different functions

Table 2 illustrates the optimization results with five different algorithms in different dimensions of solutions, where the best results among the four indices are highlighted in bold font. As seen from the table, all algorithms are useful for optimizing the benchmark functions, The proposed AgABC algorithm has the highest optimization accuracy corresponding to the index Best-F in all benchmark functions. Especially for function $f_2(x)$, if the dimensions are 60, the optimal solution Best-F is improved by 97.79% compared with the suboptimal solution, and the average fitness values Avg-F is improved by 98.87%, which means AgABC algorithm's global search ability and convergence ability are the best. On the other hand, the standard deviation *std* of AgABC algorithm is dominant in the five groups of functions, where the stability of AgABC could be proved.

From the analysis above, it can be seen that the proposed AgABC algorithm is able to produce better solutions with higher stability than any other mentioned algorithms, and the ability of exploration and exploitation of traditional algorithm could be balanced.

Table 2: The performance with different dimensions of ABC, CABC, DuABC, GABC and AgABC

| Dimensions | Functions | Algorithms | Best-F | Avg-F | std |
|---|---|---|---|---|---|
| 60 | $f_1(x)$ | ABC | 2.4845E-13 | 2.5827E-13 | 3.9760E-14 |
| | | CABC | 1.6769E-13 | 2.2496E-13 | 3.7958E-14 |
| | | DuABC | 2.7580E-05 | 2.9028E-05 | 1.0076E-10 |
| | | GABC | 1.5107E-13 | 2.0733E-13 | **2.6287E-14** |
| | | AgABC | **8.2463E-14** | **1.3554E-13** | 4.6088E-14 |
| | $f_2(x)$ | ABC | 1.3269E-11 | 2.6800E-11 | 6.4558E-12 |
| | | CABC | 1.8245E-11 | 2.6508E-11 | 5.6187E-12 |
| | | DuABC | 4.2212E+01 | 6.8358E+01 | 1.7653E-01 |
| | | GABC | 1.6781E-11 | 2.6105E-11 | 4.1657E-12 |
| | | AgABC | **2.9310E-13** | **2.9488E-13** | **1.2564E-13** |
| | $f_3(x)$ | ABC | 1.8400E-02 | 1.2600E-02 | 3.0000E-03 |
| | | CABC | 1.2000E-02 | **9.8000E-03** | 2.2000E-03 |
| | | DuABC | 9.8000E-03 | 1.4700E-02 | 6.1000E-03 |
| | | GABC | 9.9000E-03 | 1.2700E-02 | 3.1000E-03 |
| | | AgABC | **9.8000E-03** | 1.0100E-02 | **1.9000E-03** |
| | $f_4(x)$ | ABC | 1.5317E-19 | 2.2264E-19 | 4.5940E-20 |
| | | CABC | 2.1467E-19 | 3.1303E-19 | 5.0201E-20 |
| | | DuABC | 3.6004E-09 | 1.4060E-08 | 2.8505E-09 |
| | | GABC | 8.1056E-20 | 2.0281E-19 | 4.4103E-20 |
| | | AgABC | **5.9957E-20** | **1.1775E-19** | **4.1585e-20** |
| | $f_5(x)$ | ABC | 2.2210E-19 | 5.5271E-19 | 2.2210E-19 |
| | | CABC | 3.6094E-19 | 7.0393E-19 | 2.0221E-19 |
| | | DuABC | 1.7903E-10 | 5.3993E-06 | 8.8180E-06 |
| | | GABC | 3.1095E-19 | 5.2617E-19 | **1.3491E-19** |
| | | AgABC | **1.1076E-19** | **1.8253E-19** | 4.4767E-19 |
| | $f_6(x)$ | ABC | 1.1102E-16 | 1.7764E-16 | 1.1616E-16 |
| | | CABC | 0 | 1.3323E-16 | 3.0718E-16 |
| | | DuABC | 6.1530E-09 | 5.5373E-07 | 1.3563E-06 |
| | | GABC | 0 | 1.1102E-16 | 1.1947E-16 |
| | | AgABC | **0** | **1.1102E-16** | **1.1616E-16** |
| | $f_7(x)$ | ABC | 1.2250E-01 | 1.9290E-01 | 3.0800E-02 |
| | | CABC | 1.7140E-01 | 2.0260E-01 | 2.3700E-02 |
| | | DuABC | 2.6110E-01 | 3.3020E-01 | 3.2600E-02 |
| | | GABC | 1.5280E-01 | 2.0100E-01 | 2.5200E-02 |
| | | AgABC | **3.9700E-02** | **5.8900E-02** | **6.1000E-03** |
| | $f_1(x)$ | ABC | 1.5037E-07 | 8.8739E-07 | 2.2146E-08 |
| | | CABC | 1.4634E-07 | 1.7181E-07 | 1.6903E-08 |
| | | DuABC | 1.0455E+00 | 7.4861E+00 | 8.8770E+00 |
| | | GABC | 1.2074E-07 | 1.2753E-07 | 1.6523E-08 |
| | | AgABC | **1.1696E-07** | **1.1229E-07** | **1.6013E-08** |
| 80 | $f_2(x)$ | ABC | 5.7515E-07 | 7.1932E-07 | **1.1441E-07** |
| | | CABC | 4.7501E-07 | 6.8335E-07 | 1.1795E-07 |
| | | DuABC | 7.9487E+01 | 1.3056E+02 | 2.8805E+01 |
| | | GABC | 4.3346E-07 | 6.7973E-07 | 1.0741E-07 |
| | | AgABC | **5.1010E-08** | **3.9669E-07** | 3.1488E-07 |
| | $f_3(x)$ | ABC | 9.9000E-03 | **1.2900E-02** | 3.7000E-03 |
| | | CABC | 9.9000E-03 | 1.3800E-02 | 3.7000E-03 |
| | | DuABC | 9.8000E-03 | 1.6200E+02 | 6.2000E-03 |
| | | GABC | 1.0300E-02 | 1.3300E-02 | 3.6000E-03 |
| | | AgABC | **4.3336E-04** | 1.3100E-02 | **3.5300E-03** |
| | $f_4(x)$ | ABC | 8.3333E-14 | 1.1788E-13 | **1.9037E-14** |
| | | CABC | 9.6407E-14 | 1.1713E-13 | 1.7885E-14 |
| | | DuABC | 1.5636E-05 | 1.3000E-03 | 2.0000E-03 |
| | | GABC | 8.9666E-14 | 1.1268E-13 | 2.0044E-14 |
| | | AgABC | **7.3118E-14** | **9.6293E-14** | 2.0508E-14 |
| | $f_5(x)$ | ABC | 1.2891E-13 | 2.8900E-13 | 1.0582E-13 |
| | | CABC | 2.0006E-13 | 3.1890E-13 | 6.8818E-14 |
| | | DuABC | 1.1404E-06 | 9.1349E-04 | 1.6000E-03 |
| | | GABC | 1.7714E-13 | 2.4924E-13 | **5.4782E-14** |
| | | AgABC | **9.0123E-14** | **1.6603E-13** | 8.2258E-14 |
| | $f_6(x)$ | ABC | 1.7208E-14 | 2.5074E-14 | 1.1272E-14 |
| | | CABC | 1.7097E-14 | 2.3870E-14 | 7.3142E-15 |
| | | DuABC | 2.4041E-08 | 6.0038E-05 | 8.6160E-05 |
| | | GABC | 1.6653E-14 | 2.1427E-14 | 6.0335E-15 |
| | | AgABC | **1.5654E-14** | **2.0280E-14** | **4.6634E-15** |
| | $f_7(x)$ | ABC | 1.4850E-01 | 2.2630E-01 | 3.1500E-02 |
| | | CABC | 1.8270E-01 | 2.1750E-01 | 2.4100E-02 |
| | | DuABC | 2.6630E-01 | 3.4490E-01 | 3.6400E-02 |
| | | GABC | 1.5820E-01 | 2.1880E-01 | **2.7800E-03** |
| | | AgABC | **5.9900E-02** | **7.3800E-02** | 1.3200E-02 |
| | $f_1(x)$ | ABC | 2.1839E-04 | 3.4023E-04 | 5.2402E-05 |
| | | CABC | 2.6330E-04 | 3.3888E-04 | 4.6597E-05 |
| | | DuABC | 1.7158E+01 | 2.0665E+03 | 3.4359E+03 |
| | | GABC | 2.9942E-04 | 3.4304E-04 | **3.2313E-05** |
| | | AgABC | **1.2351E-04** | **2.1673E-04** | 8.0907E-05 |
| | $f_2(x)$ | ABC | 1.8246E-04 | 2.3876E-04 | 3.4522E-05 |
| | | CABC | 1.2879E-04 | **2.2308E-04** | **3.4283E-05** |
| | | DuABC | 1.3422E+02 | 1.8732E+02 | 3.5678E+01 |
| | | GABC | 1.6861E-04 | 2.2445E-04 | 3.9883E-05 |
| | | AgABC | **3.1639E-05** | 3.8098E-04 | 4.9405E-04 |
| | $f_3(x)$ | ABC | 6.0000E-03 | 1.3100E-02 | 3.9000E-03 |
| | | CABC | 9.8000E-03 | 1.3200E-02 | 3.9000E-03 |
| | | DuABC | 1.0000E-02 | 1.3100E-02 | 3.4000E-03 |
| | | GABC | 9.9000E-03 | 1.4400E-02 | 4.4000E-03 |
| | | AgABC | **6.3567E-04** | **3.3100E-03** | **3.3300E-03** |
| | | ABC | 2.4860E-10 | 3.2472E-10 | 4.7603E-11 |
| | | CABC | 2.6839E-10 | 3.6351E-10 | 5.6069E-11 |
| 100 | $f_4(x)$ | DuABC | 6.4436E-05 | 8.3000E-03 | 1.0800E-02 |
| | | GABC | 2.4771E-10 | 3.2844E-10 | 4.2017E-11 |
| | | AgABC | **1.3110E-11** | **1.6913E-11** | **5.3728E-12** |
| | | ABC | 4.0892E-10 | 7.0459E-10 | 2.3648E-10 |
| | | CABC | 5.3481E-10 | 8.6121E-10 | 1.9300E-10 |
| | $f_5(x)$ | DuABC | 4.8342E-05 | 1.5400E-02 | 3.2800E-02 |
| | | GABC | 3.6065E-10 | 7.5293E-10 | 2.1563E-10 |
| | | AgABC | **2.9560E-11** | **4.3616E-11** | **1.9021E-11** |
| | | ABC | 1.3461E-11 | 2.3187E-11 | 7.3313E-12 |
| | | CABC | 1.2709E-11 | 2.6427E-11 | 9.2797E-12 |
| | $f_6(x)$ | DuABC | 1.5931E-06 | 3.9960E-04 | 9.4855E-04 |
| | | GABC | 1.2323E-11 | 2.2055E-11 | 5.4949E-12 |
| | | AgABC | **2.2913E-12** | **2.6990E-12** | **5.7654E-13** |
| | | ABC | 1.6950E-01 | 2.3230E-01 | 3.0300E-02 |
| | | CABC | 1.5170E-01 | 2.2470E-01 | 2.7900E-02 |
| | $f_7(x)$ | DuABC | 2.9950E-01 | 3.7080E-01 | 3.7500E-02 |
| | | GABC | 1.6980E-01 | 2.2900E-01 | 3.4400E-02 |
| | | AgABC | **5.1400E-02** | **6.5400E-02** | **9.4767E-03** |

## Simulation of task set scheduling problem

After verifying the performance of the proposed algorithms on complex mathematical problems, in this section it is applied to solve the scheduling problem in the container storage area0 where the optimal solution is the sequence of inbound and outbound tasks with twin ETVs, and the integer encoding scheme is introduced to represent different sequence. This scheme assigns random numbers between -10 and 10 to each dimension of the generated solution, and sorts them in ascending order based on the values of the random numbers, the sequence generated by the index values is the corresponding scheduling scheme.

There are 16 entrances and exits in the container storage area of the airport station, the entrance coordinates are $R_1$ (1-1-5), $R_2$ (1-1-15), $R_3$ (2-1-20), $R_4$ (1-1-25), $R_5$ (1-1-30), $R_6$ (1-1-35), $R_7$ (1-1-40), $R_8$ (1-1-50) and $R_9$ (1-1-60), and the exit coordinates are $C_1$(1-1-8), $C_2$(1-1-18), $C_3$(1-1-28), $C_4$(1-1-38), $C_5$(2-1-48), $C_6$(1-1-53) and $C_7$(1-1-58), where $R_3$ and $C_5$ are I/O ports at the landside, while the rest are I/O ports at the airside. Here, the first value in the bracket of coordinates represents the number of row of a shelf, the second value indicates the number of the layer and the third value is the number of the column. Table 3 depicts the assigned or current position of 60 tasks needed to be scheduled, where the first 30 ones are input tasks, and the last 30 ones are output tasks.

A comparative study among different intelligent algorithms, including ABC, PSO0 and AgABC, CABC, GABC and DuABC, are executed for the above scheduling problem. The swarm size of all algorithms is set to 200 with 60 dimensions, the maximum local search time is 50, the stopping criterion is set to 1000 generations. Initial populations are generated by uniformly random sampling from the searching space in all algorithms considered. Each algorithm runs 20 trials independently.

Table 3: Assigned positions of inbound or outbound tasks

|    | Assigned position of inbound tasks |    | Current position of outbound tasks |    | Assigned position of inbound tasks |    | Current position of outbound tasks |
|----|----|----|----|----|----|----|----|
| 1  | I(1-5-10) | 31 | O(1-3-10) | 16 | I(1-8-44) | 46 | O(2-8-10) |
| 2  | I(2-3-14) | 32 | O(1-5-55) | 17 | I(2-8-32) | 47 | O(1-3-32) |
| 3  | I(1-3-23) | 33 | O(1-5-25) | 18 | I(2-3-54) | 48 | O(1-4-50) |
| 4  | I(1-5-26) | 34 | O(2-4-8)  | 19 | I(1-3-40) | 49 | O(2-3-38) |
| 5  | I(1-5-30) | 35 | O(2-2-18) | 20 | I(1-4-60) | 50 | O(2-1-58) |
| 6  | I(1-2-18) | 36 | O(2-1-16) | 21 | I(1-3-20) | 51 | O(1-5-24) |
| 7  | I(1-5-24) | 37 | O(2-3-51) | 22 | I(2-2-43) | 52 | O(1-4-30) |
| 8  | I(1-4-40) | 38 | O(1-5-6)  | 23 | I(2-4-50) | 53 | O(2-6-40) |
| 9  | I(1-5-40) | 39 | O(2-5-3)  | 24 | I(1-6-10) | 54 | O(2-4-35) |
| 10 | I(1-5-35) | 40 | O(1-6-12) | 25 | I(2-7-20) | 55 | O(2-8-51) |
| 11 | I(2-5-23) | 41 | O(2-6-13) | 26 | I(1-6-15) | 56 | O(2-2-30) |
| 12 | I(1-7-43) | 42 | O(2-7-49) | 27 | I(2-8-30) | 57 | O(1-2-60) |
| 13 | I(1-3-48) | 43 | O(1-7-57) | 28 | I(2-2-45) | 58 | O(1-3-26) |
| 14 | I(1-8-50) | 44 | O(1-5-25) | 29 | I(1-7-58) | 59 | O(1-6-35) |
| 15 | I(1-6-21) | 45 | O(2-6-18) | 30 | I(1-4-9)  | 60 | O(2-8-45) |

Table 4 presents the optimal fitness values and their average values with different algorithms in the first ten trials. It can be seen from the above analysis, the proposed AgABC algorithm possesses the shortest average scheduling time in six different algorithms, and it obtained the best fitness value in all ten trials, which could be improved by 2.8% compared with the worst one.

The scheduling sequence or assignment with AgABC is listed in Table 5. Obviously the six algorithms all can handle the complicated task set scheduling problem with constraints, and compared with other algorithms, AgABC can improve the operation efficiency effectively.

Figure 6 shows the sequence of task set optimized with AgABC algorithm and the corresponding trajectories of two ETVs, respectively. It is clear that the twin ETVs could execute the scheduling tasks successfully without conflicting with each other.

Table 4: The scheduling results with different algorithms

|       | 1 | 2 | 3 | 4 | 5 | 6 | 7 | 8 | 9 | 10 | Avg(s) |
|-------|---|---|---|---|---|---|---|---|---|----|--------|
| ABC   | 2793.17 | 2801.13 | 2793.17 | 2783.3  | 2784.07 | 2793.07 | 2789.08 | 2783.3  | 2801.13 | 2784.07 | 2790.549 |
| CABC  | 2737.44 | 2742.29 | 2741.68 | 2737.44 | 2784.07 | 2737.44 | 2742.29 | 2737.44 | 2784.07 | 2737.44 | 2748.221 |
| DuABC | 2742.51 | 2752.74 | 2793.07 | 2742.51 | 2752.74 | 2783.3  | 2742.51 | 2742.51 | 2793.07 | 2742.51 | 2758.747 |
| GABC  | 2733.96 | 2748.81 | 2733.96 | 2748.81 | 2733.96 | 2733.96 | 2742.51 | 2748.81 | 2733.96 | 2733.96 | 2739.27  |
| AgABC | 2739.32 | 2728.48 | 2739.32 | 2750.13 | 2732.84 | 2739.32 | 2732.84 | 2732.84 | 2739.32 | 2728.48 | 2736.289 |
| PSO   | 2796.26 | 2795.73 | 2796.26 | 2795.73 | 2796.26 | 2796.26 | 2796.26 | 2795.73 | 2795.73 | 2796.26 | 2796.048 |

Table 5: Entrance and exit allocation scheme corresponding to AgABC

|   | Input |   | Output |
|---|-------|---|--------|
| 1 | 2, 21, 27     | 1 | 33, 50, 47, 58, 40, 43, 48 |
| 2 | 11, 28        | 2 | 36, 60, 54, 31 |
| 3 | 6, 26, 24, 19 | 3 | 45, 38, 32, 56, 59, 39 |
| 4 | 5, 4, 9, 20   | 4 | 49, 41, 57, 52 |
| 5 | 3, 16, 30     | 5 | 42, 46, 51, |
| 6 | 10, 22,       | 6 | 35, 37, 53 |
| 7 | 23, 7, 12, 17, 25, 15 | 7 | 34, 44, 55 |
| 8 | 13, 29, 18, 8 | | |
| 9 | 1, 14 | | |

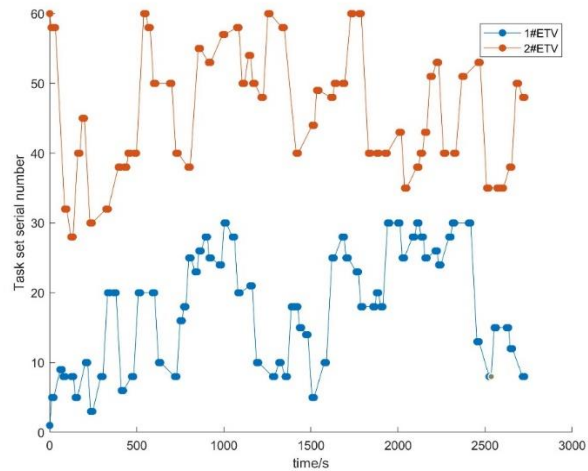

Figure 6: The trajectories of double ETVs

From the analysis above, it is believed that the artificial bee colony algorithm based on adaptive group collaborative mode increases the depth of exploration and expands the search range.

## Conclusions

An adaptive group collaborative ABC algorithm, whose update process is guided by the group collaboration mechanism, is introduced in this paper. Different search strategies with different functionality are assigned to different members in each group, and the probability to be selected is adjusted dynamically during the optimization procedure. The performance of AgABC has been verified to be superior to other improved ABC algorithms on several benchmark functions, and it can effectively balance the ability of exploration and exploitation. The computational experiment was carried out on task set scheduling problem, it could effectively avoid conflicts and generate valid scheduling sequences. To conclude, the proposed framework provides more promising solutions to guide the evolution and effectively helps ABC algorithms escaping the situation of local optimal.

## Data Availability

No data were used to support this study.

## Conflicts of Interest

The authors declare that there are no conflicts of interest regarding the publication of this paper.

## Funding Statement

This work was supported by the Training Program for Young Teachers in Universities of Henan Province[grant number 2020GGJS137]; the NSFC-Zhejiang Joint Fund for the Integration of Industrialization and Informatization[grant number U1709215]; the Zhejiang

Province Key R&D projects,[grant number No.2019C03104]; the Fundamental Research Funds of Zhongyuan University of Technology[grant number K2019YY005]; Henan Province Science and Technology R&D projects[grant number 20202102210135, 212102310547 and 212102210080]; National High-level Foreign Expert Program[grant number G2021026006L].

## References

[1] Karaboga D and Basturk B, "Artificial Bee Colony(ABC) optimization algorithm for solving constrained optimization problems," *The 12th International Fuzzy Systems Association World Congress*, pp. 789-798, 2007.

[2] Haiquan Wang, Jianhua Wei, Shengjun Wen et al., "Improved Artificial Bee Colony Algorithm and its Application in Classification," *Journal of Robotics and Mechatronics*, vol. 30, no. 6, pp. 921-926, 2018.

[3] Devi M K and Umamaheswari K, "Modified Artificial Bee Colony with firefly algorithm based spectrum handoff in cognitive radio network," *International Journal of Intelligent Networks*, vol. 1, pp. 67–75, 2020.

[4] Rambabu B, Reddy A V and Janakiraman S, "Hybrid Artificial Bee Colony and Monarchy Butterfly Optimization Algorithm (HABC-MBOA)-based cluster head selection for WSNs," *Journal of King Saud University - Computer and Information Sciences*, 2019.

[5] Zhenping Liang, Kaifeng Hu, Quanxiang Zhu et al., "An enhanced artificial bee colony algorithm with adaptive differential operators," *Applied Soft Computing Journal*, vol. 58, pp. 480–494, 2017.

[6] Xu Chen, Xuan Wei, Guanxue Yang et al., "Fireworks explosion based artificial bee colony for numerical optimization," *Knowledge-Based Systems*, vol. 188, Article ID 105002, 2020.

[7] Xiaoyu Song, Ming Zhao, Qifeng Yan et al., "A high-efficiency adaptive artificial bee colony algorithm using two strategies for continuous optimization," *Swarm and Evolutionary Computation*, vol. 50, Article ID 100549, 2019.

[8] Feiyi Xu, Haolun Li, Chi-Man Pun et al., "A new global best guided artificial bee colony algorithm with application in robot path planning," *Applied Soft Computing Journal*, vol. 88, Article ID 106037, 2020.

[9] LEI De ming and YANG Hai, "Multi-colony artificial bee colony algorithm for multi-objective unrelated parallel machine scheduling problem," *Control and Decision*, vol. 2021, no. 4, pp. 1-9, 2021.

[10] Weifeng Gao, Zhifang Wei, Yuting Luo et al., "Artificial bee colony algorithm based on parzen window method," *Applied Soft Computing Journal*, vol. 74, pp. 679–692, 2019.

[11] Singh K, Sundar S. "Artifical bee colony algorithm using problem-specific neighborhood strategies for the tree t-spanner problem," *Applied Soft Computing*, vol.62, pp:110-118,2018.

[12] Charin Chanuri, Ishak Dahaman, Mohd Zainuri Muhammad Ammirrul Atiqi et al., "Modified Levy Flight Optimization for a Maximum Power Point Tracking Algorithm under Partial Shading," *Applied Sciences*, vol. 11, no. 3, pp. 992, 2021.

[13] Wang Y, Guo G D and Chen L F, "Chaotic Artificial Bee Colony algorithm: A new approach to the problem of minimization of energy of the 3D protein structure," *Molecular Biology*, vol. 47, no. 6, pp. 894-900, 2013.

[14] Guopu Zhu, Sam Kwong, "Gbest-guided artificial bee colony algorithm for numerical function optimization," *Applied Mathematics and Computation*, Vol.217, No.2010, pp.3166-3173,2010

[15] Turky A, Abdullah S and Dawod A, "A dual-population multi operators harmony search algorithm for dynamic optimization problems," *Computers & Industrial Engineering*, vol. 117, pp. 19-28, 2018.

[16] Haiquan Wang, Jianhua Wei, Menghao Su et al., "Task Set Scheduling of Airport Freight Station Based on Parallel Artificial Bee Colony Algorithm," *14th International Conference on Bio-inspired Computing: Theories and Applications (BIC-TA)*, Zhengzhou, 2020.

[17] Loau Tawfak Al-Bahrani, Jagdish C. Patra, "A novel orthogonal PSO algorithm based on orthogonal diagonalization," *Swarm and Evolutionary Computation*, vol. 40, pp. 1-23, 2018.